\pdfoutput=1

\documentclass[11pt]{article}

\usepackage[]{acl2024}
\usepackage{amssymb}
\usepackage{ragged2e}
\usepackage{times}
\usepackage{latexsym}
\usepackage{pgfplots}
\usepackage{mathrsfs}
\usepackage{array}
\newcolumntype{P}[1]{>{\centering\arraybackslash}p{#1}}
\usepackage{supertabular}
\usepackage{longtable}
\pgfplotsset{compat=1.18, width=7.7cm}
\usepackage[T1]{fontenc}
\usepackage[utf8]{inputenc}
\usepackage[table]{xcolor}

\usepackage{subfig}
\usepackage{microtype}
\usepackage{multirow}
\usepackage{graphicx}

\usepackage{booktabs} 
\usepackage{algpseudocode}
\usepackage{tcolorbox}
\tcbuselibrary{breakable} 
\usepackage{lipsum}
\usepackage{listings}
\newtcolorbox{codebox}[1][]{colback=white,
colframe=blue!75!black,
fonttitle=\bfseries,
title= Example,
  breakable,
  #1}
  \newtcolorbox{querybox}[1][]{
  colback=red!5!white,
  colframe=red!75!black,
  fonttitle=\bfseries,
  title=query:,
  breakable,
  #1
}
  \newtcolorbox{cbox}[1][]{
  colback=green!5!white,
  colframe=green!75!black,
  fonttitle=\bfseries,
  title=query:,
  breakable,
  #1
}
  \newtcolorbox{abox}[1][]{
  colback=yellow!5!white,
  colframe=yellow!75!black,
  fonttitle=\bfseries,
  title=query:,
  breakable,
  #1
}
\usepackage{hyperref}

\usepackage{amsmath}

\usepackage{algorithm}
\usepackage{algpseudocode}

\AtBeginDocument{%
  }

\title{NC2C: Automated Convexification of Generic Non-Convex Optimization Problems}

\author{
~~Xinyue Peng$^{1}$
~~Yanming Liu$^{2}$
~~Yihan Cang$^{1}$
~~Yuwei Zhang \\ \bf
~~Xinyi Wang
~~Songhang Deng
~~Jiannan Cao$^{3}$
 \\
$^{1}$Southeast University $^{2}$Zhejiang University \\ 
$^{3}$Massachusetts Institute of Technology \\
\texttt{{xinyuepeng@seu.edu.cn, oceann24@zju.edu.cn}}\\
\texttt{{jiannan@mit.edu}}\\
}

\begin{document}

\maketitle

\begin{abstract}
Non-convex optimization problems are pervasive across mathematical programming, engineering design, and scientific computing, often posing intractable challenges for traditional solvers due to their complex objective functions and constrained landscapes. To address the inefficiency of manual convexification and the over-reliance on expert knowledge, we propose NC2C, an LLM-based end-to-end automated framework designed to transform generic non-convex optimization problems into solvable convex forms using large language models. NC2C leverages LLMs' mathematical reasoning capabilities to autonomously detect non-convex components, select optimal convexification strategies, and generate rigorous convex equivalents. The framework integrates symbolic reasoning, adaptive transformation techniques, and iterative validation, equipped with error correction loops and feasibility domain correction mechanisms to ensure the robustness and validity of transformed problems. Experimental results on a diverse dataset of 100 generic non-convex problems demonstrate that NC2C achieves an 89.3\% execution rate and a 76\% success rate in producing feasible, high-quality convex transformations. This outperforms baseline methods by a significant margin, highlighting NC2C's ability to leverage LLMs for automated non-convex to convex transformation, reduce expert dependency, and enable efficient deployment of convex solvers for previously intractable optimization tasks.
\end{abstract}

\section{Introduction}
Non-convex optimization problems are ubiquitous across various domains, including mathematical programming, engineering design, machine learning, and scientific computing \cite{sui2024non, mikhalevich2024methods}.  Convex optimization offers stable and efficient solutions, especially when closed-form solutions are available \cite{garatti2025non}. However, many real-world optimization problems are inherently non-convex, characterized by complex objective functions, non-linear constraints, and coupled variables that create multiple local optima \cite{liu2024quantum}. Addressing these challenges often requires sophisticated modeling, transformation, and relaxation techniques to convert non-convex problems into convex ones, typically depending heavily on expert knowledge in advanced mathematical optimization, problem-specific domain expertise, and transformation strategies \cite{jiang2024origins}.

To reduce reliance on expert knowledge, recent research shows that large language models (LLMs) show significant potential in tackling complex mathematical problems. For instance, LLMs achieve 97.1\% accuracy in solving mathematical challenges with zero-shot prompting on the GSM8K dataset \cite{zhong2024achieving}. In a series of reasoning and derivations involving complex mathematical proof lemmas, a wide range of large language model–based approaches, such as Seed-Prover \cite{chen2025seed, chen2025seed2}, have demonstrated the strong capability of LLMs in relational reasoning and lemma-proof inference. They can also address classical optimization problems, such as linear regression and the traveling salesman problem, through iterative methods \cite{yang2024large}. Moreover, LLMs can leverage existing solvers like \texttt{Gurobi} and \texttt{CVXPY} for optimization tasks \cite{b2}. For example, the \textit{gurobi.abs} function enables direct modeling of $\ell_1$-norm objectives, allowing LLMs to avoid redundant auxiliary constraints and variables, thus accelerating the solving process \cite{ahmaditeshnizi2023optimus}. However, these solvers primarily handle convex problems, limiting their ability to solve non-convex challenges directly,  so LLMs must draw on their own capabilities to transform and solve non-convex challenges.

To address the existing gap in applying LLMs to automated non-convex to convex transformation, we introduce NC2C, an LLM-based framework that leverages advancements in mathematical reasoning, prompt engineering, and few-shot learning. The main contributions of this paper are summarized as follows.
\begin{itemize}
    \item We propose the NC2C framework, an LLM-based approach to address generic non-convex optimization problems. The framework automatically detects non-convex components and efficiently transforms them into convex problems, significantly reducing reliance on expert knowledge and enabling automated non-convex problem solving across diverse domains.
    \item We introduce Error Correction Loops (ECL) and Feasibility Domain Correction (FDC) as refinement stages within the NC2C framework. These enhancements not only boost the reliability and performance of NC2C's solutions but also improve other existing advanced methods, demonstrating the framework's robustness in handling complex optimization transformations.
    \item Experimental results show that NC2C achieves an execution rate of {89.3\%} and a success rate of {76\%} on the GPT-4 model, significantly outperforming baseline methods. These results highlight NC2C's effectiveness and robustness in solving generic non-convex optimization problems through automated convexification.
\end{itemize}

\begin{figure*}[htbp]
    \centering
    \includegraphics[width=0.8\textwidth]{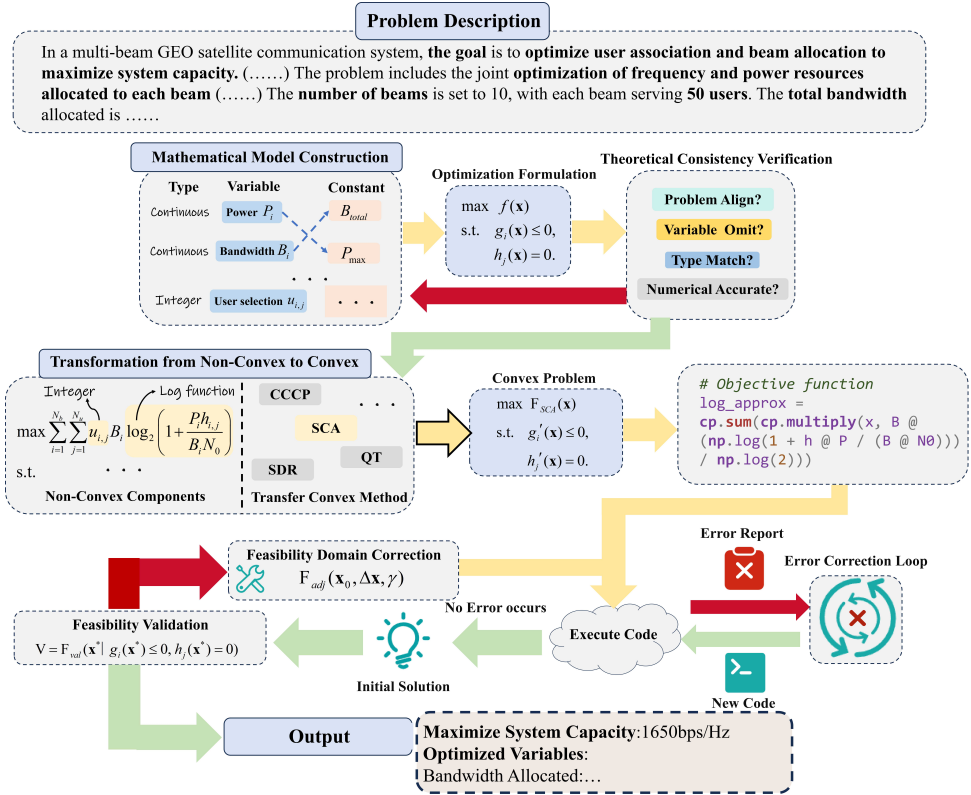}
    \caption{Overview of the NC2C framework, which leverages LLMs to automatically transform generic non-convex optimization problems into convex forms. }
    \label{fig:main}
\end{figure*}

\section{Related Work}
\label{others}

\subsection{LLMs for Optimization Problems.} Recent studies have shown that large language models (LLMs) have significant potential in addressing complex optimization problems, both convex and non-convex. LLMs have demonstrated success in solving classical optimization problems like linear regression and the traveling salesman problem using iterative methods~\citep{yang2024large}. Moreover, LLMs can leverage existing solvers like \texttt{Gurobi} and \texttt{CVXPY} to handle convex optimization tasks, accelerating the solving process~\citep{ahmaditeshnizi2023optimus}. LLMOPT \cite{jiang2025llmopt} and OptimAI \cite{thind2025optimai} focus on end-to-end frameworks that define and solve problems from scratch using autonomous agents. To enhance reliability, OR-LLM-Agent \cite{zhang2025or} leverages reasoning-heavy models to automate complex operations research modeling. Meanwhile, OptiBench \cite{yangoptibench} introduces standardized benchmarks to systematically measure and improve LLMs' modeling accuracy. However, these solvers primarily focus on convex problems, and LLMs must utilize their inherent capabilities to transform and solve non-convex problems, as direct solutions remain a challenge~\citep{zhong2024achieving}.

\subsection{Tool Learning and Reasoning for LLMs.} 

The integration of LLMs with external tools represents a paradigm shift from closed-loop generation to open-environment interaction. Beyond basic function calling \cite{instructgpt}, frameworks like ReAct\cite{yao2022react} and ToolChain*\cite{zhuangtoolchain} formalize the interleaving of reasoning and acting. To enhance long-horizon consistency, Tool-Planner \cite{qintoolllm, liu2025toolplanner} introduces explicit global planning, transitioning from reactive execution to proactive strategy optimization.

To overcome the limitations of linear decision-making, LATS \cite{zhou2024language} incorporates tree-search heuristics to traverse the strategy space, while Reflexion enables iterative self-correction through feedback loops, establishing a closed-loop refinement mechanism for tool invocation \cite{schick2023toolformer}. Consequently, benchmarks \cite{ye2025tooleyes, patil2024gorilla} have evolved from evaluating API selection accuracy to assessing high-order dimensions . These paradigms are now permeating multimodal GUI Agents \cite{zhang2025api} and specialized domains, with Reinforcement Learning \cite{qian2025toolrl} further aligning dynamic reasoning policies. 
\section{Methodology}
As illustrated in Figure \ref{fig:main}, we propose the NC2C framework, which leverages LLMs to automatically transform generic non-convex optimization problems into convex forms.

\subsection{Problem Formulation and Mathematical Modeling}\label{math} 
Given a natural language problem description \(\mathcal{D}_p\) provided by the user, NC2C first constructs a formal mathematical representation. The framework employs Named Entity Recognition (NER) methods \cite{wang2023gpt} to extract optimization-related entities from \(\mathcal{D}_p\), identifying decision variables, parameters, and their relationships.

Let \(\mathcal{V} = \{v_1, v_2, \ldots, v_n\}\) denote the set of extracted variables, where each variable \(v_i\) is associated with a type \(\tau_i \in \{\text{continuous}, \text{integer}, \text{binary}\}\). The variable extraction process is formalized as:

\begin{equation}
    \mathcal{V} = \text{ExtractVariables}(\mathcal{D}_p)
\end{equation}

\begin{equation}
    \tau_i = \text{InferType}(v_i, \mathcal{D}_p).
\end{equation}

The objective function is identified by analyzing optimization goals expressed in \(\mathcal{D}_p\). The framework constructs the objective function \(f: \mathbb{R}^n \rightarrow \mathbb{R}\) by mapping extracted variables to their mathematical relationships. Constraint extraction identifies inequality constraints \(g_i(\mathbf{x}) \leq 0\) and equality constraints \(h_j(\mathbf{x}) = 0\) from the problem description.

The complete optimization problem \(\mathcal{P}\) is formulated as:
\begin{equation}
\begin{aligned}
\quad \underset{\mathbf{x} \in \mathcal{X}}{\text{optimize}} \quad &f(\mathbf{x}) \\
\text{subject to} \quad &g_i(\mathbf{x}) \leq 0, \quad i \in \mathcal{I} = \{1, 2, \ldots, m\}, \\
&h_j(\mathbf{x}) = 0, \quad j \in \mathcal{J} = \{1, 2, \ldots, p\}, \\
&\mathbf{x} \in \mathcal{X},
\end{aligned}
\end{equation}
where \(\mathbf{x} = [x_1, x_2, \ldots, x_n]^T\) is the decision vector, \(\mathcal{X}\) denotes the variable domain constraints, \(m\) and \(p\) represent the number of inequality and equality constraints, respectively, and the optimization direction (minimize or maximize) is determined from \(\mathcal{D}_p\). 

\subsection{Non-Convex Detection and Convexification}\label{convex}

NC2C systematically identifies non-convex components in \(\mathcal{P}\) and applies appropriate convexification strategies. The non-convex detection process analyzes the structure of \(f(\mathbf{x})\), \(g_i(\mathbf{x})\), and \(h_j(\mathbf{x})\) to identify problematic terms such as bilinear products, fractional expressions, logarithmic functions in maximization contexts, or integer constraints.

Let \(\mathcal{N} = \{n_1, n_2, \ldots, n_k\}\) denote the set of identified non-convex components. For each component \(n_l \in \mathcal{N}\), the framework selects a convexification strategy \(\sigma_l\) from a strategy set \(\Sigma = \{\text{SCA}, \text{SDR}, \text{Lagrangian}, \text{Substitution}, \ldots\}\). The strategy selection is formalized as:
\begin{equation}
    \sigma_l = \text{SelectStrategy}(n_l, \mathcal{P}), \quad \forall n_l \in \mathcal{N}.
\end{equation}

For non-convex objective functions, NC2C commonly employs Successive Convex Approximation (SCA). Given a reference point \(\mathbf{x}^{(0)} \in \mathcal{X}\), SCA constructs a convex approximation \(\tilde{f}(\mathbf{x}; \mathbf{x}^{(0)})\) using first-order Taylor expansion:
\begin{equation}
    \tilde{f}(\mathbf{x}; \mathbf{x}^{(0)}) = f(\mathbf{x}^{(0)}) + \nabla f(\mathbf{x}^{(0)})^T (\mathbf{x} - \mathbf{x}^{(0)}),
\end{equation}
where \(\nabla f(\mathbf{x}^{(0)})\) denotes the gradient of \(f\) evaluated at \(\mathbf{x}^{(0)}\). The convexified objective function is then:
\begin{equation}
    f_c(\mathbf{x}) = \text{Convexify}(f(\mathbf{x}), \{\sigma_l\}_{l=1}^k, \mathbf{x}^{(0)}).
\end{equation}

Non-convex constraints are similarly transformed. For inequality constraints \(g_i(\mathbf{x}) \leq 0\), the convexified version becomes:
\begin{equation}
    \tilde{g}_i(\mathbf{x}; \mathbf{x}^{(0)}) = g_i(\mathbf{x}^{(0)}) + \nabla g_i(\mathbf{x}^{(0)})^T (\mathbf{x} - \mathbf{x}^{(0)}) \leq 0,
\end{equation}
resulting in the convexified problem:
\begin{equation}
\begin{aligned}
\mathcal{P}_c: \quad \underset{\mathbf{x} \in \mathcal{X}}{\text{optimize}} \quad &f_c(\mathbf{x}) \\
\text{subject to} \quad &\tilde{g}_i(\mathbf{x}; \mathbf{x}^{(0)}) \leq 0, \quad i \in \mathcal{I}, \\
&\tilde{h}_j(\mathbf{x}) = 0, \quad j \in \mathcal{J}, \\
&\mathbf{x} \in \mathcal{X}_c,
\end{aligned}
\end{equation}
where \(\mathcal{X}_c\) represents the convexified variable domain.

\subsection{Code Generation and Execution}
Based on the convexified problem \(\mathcal{P}_c\), NC2C generates executable Python code \(C(\mathcal{P}_c)\) that implements the optimization model using appropriate solvers. The framework selects a solver \(\mathcal{S} \in \{\text{CVXPY}, \text{Gurobi}, \text{SCIPY}\}\) based on problem characteristics: \texttt{CVXPY}\cite{b2} for standard convex optimization, \texttt{Gurobi} for large-scale linear and mixed-integer problems, and \texttt{SCIPY} for general nonlinear convex problems.

The code generation process is formalized as:
\begin{equation}
    C(\mathcal{P}_c) = \text{GenerateCode}(\mathcal{P}_c, \mathcal{S}).
\end{equation}

Upon execution, the code either succeeds and returns a solution \(\mathbf{x}^*\), or fails and generates an error report \(\mathcal{E}\). The execution status is captured by an indicator function:
\begin{equation}
    \mathbb{I}_e(C(\mathcal{P}_c)) = \begin{cases}
        1 & \text{if execution succeeds}, \\
        0 & \text{if execution fails}.
    \end{cases}
\end{equation}

When \(\mathbb{I}_e = 1\), the solver returns the optimal solution:
\begin{equation}
    \mathbf{x}^* = \underset{\mathbf{x} \in \mathcal{X}_c}{\arg\min} \ f_c(\mathbf{x}) \quad \text{s.t.} \ \tilde{g}_i(\mathbf{x}) \leq 0, \ \tilde{h}_j(\mathbf{x}) = 0.
\end{equation}

\subsection{Error Correction Loop (ECL)}
When code execution fails (\(\mathbb{I}_e = 0\)), NC2C activates the Error Correction Loop (ECL) to iteratively refine the generated code. In the \(k\)-th iteration (\(k = 1, 2, \ldots, K\)), ECL analyzes the error report \(\mathcal{E}^{(k)}\) to identify failure causes, such as syntax errors, dimension mismatches, or solver incompatibilities.

The error analysis and code correction process is formalized as:
\begin{equation}
    C^{(k+1)}(\mathcal{P}_c) = \text{CorrectCode}(C^{(k)}(\mathcal{P}_c), \mathcal{E}^{(k)}, \mathcal{S}, \mathcal{P}_c),
\end{equation}
where \(C^{(k)}\) denotes the code at iteration \(k\), and \(\text{CorrectCode}(\cdot)\) represents the LLM-based correction function that modifies the code based on error analysis.

The ECL process terminates when either execution succeeds or the maximum iteration count \(K\) is reached:
\begin{equation}
    \mathbf{x}^* = \begin{cases}
        \text{Solve}(C^{(k)}(\mathcal{P}_c), \mathcal{S}) & \text{if } \mathbb{I}_e(C^{(k)}(\mathcal{P}_c)), \\
        \text{ECL}(C^{(1)}(\mathcal{P}_c), \mathcal{E}, \mathcal{S}, K) & \text{otherwise},
    \end{cases}
\end{equation}
where \(\text{ECL}(\cdot)\) represents the iterative correction process that attempts up to \(K\) iterations to produce executable code.

\subsection{Solution Validation}
After obtaining a solution \(\mathbf{x}^*\) from the convexified problem \(\mathcal{P}_c\), NC2C validates its feasibility with respect to the original problem \(\mathcal{P}\). This validation is critical because convexification may relax constraints, potentially yielding solutions that violate the original constraints.

The feasibility validation checks $\forall i,j \in \mathcal{I},\mathcal{J}$ whether \(\mathbf{x}^*\) satisfies all original constraints:
\begin{equation}
    \mathbb{I}_f(\mathbf{x}^*) = \begin{cases}
        1 & \text{if } g_i(\mathbf{x}^*) \leq 0 \text{ and } h_j(\mathbf{x}^*) = 0 \ , \\
        0 & \text{otherwise}.
    \end{cases}
\end{equation}

Additionally, NC2C performs theoretical consistency validation during the mathematical modeling stage to ensure the constructed problem \(\mathcal{P}\) accurately reflects the original problem description \(\mathcal{D}_p\). This validation checks four criteria: (1) alignment between formulas and problem description, (2) completeness of variables and constraints, (3) correctness of variable types, and (4) accuracy of numerical values. The consistency metric is:
\begin{equation}
    \mathbb{I}_c(\mathcal{P}, \mathcal{D}_p) = \prod_{i=1}^{4} \mathbb{H}(\xi_i),
\end{equation}
where \(\xi_i\) represents the \(i\)-th validation criterion and \(\mathbb{H}(\cdot)\) is an indicator function returning 1 if the criterion is satisfied and 0 otherwise.

\subsection{Feasibility Domain Correction (FDC)}
When the solution \(\mathbf{x}^*\) fails feasibility validation (\(\mathbb{I}_f(\mathbf{x}^*) = 0\)), NC2C activates the Feasibility Domain Correction (FDC) mechanism to iteratively refine the solution. FDC operates in two stages over a maximum of \(L\) iterations.

\begin{itemize}
\item \textbf{Stage 1: Initial Value Adjustment} (iterations \(l = 1, 2, \ldots, \lfloor L/2 \rfloor\)): FDC adjusts the initial point \(\mathbf{x}_0\) using a correction function that gradually moves toward the feasible region:
\begin{equation}
    \mathbf{x}_0^{(l+1)} = \mathbf{x}_0^{(l)} + \alpha^{(l)} \cdot \Delta\mathbf{x}^{(l)},
\end{equation}
where \(\alpha^{(l)} \in (0, 1]\) is the step size at iteration \(l\), and \(\Delta\mathbf{x}^{(l)}\) is the correction direction computed based on constraint violations. The updated initial point is used to re-solve \(\mathcal{P}_c\), yielding a new candidate solution \(\mathbf{x}^{*(l+1)}\).

\item \textbf{Stage 2: Problem Re-convexification} (iterations \(l = \lfloor L/2 \rfloor + 1, \ldots, L\)): If Stage 1 fails to find a feasible solution, FDC reanalyzes the original problem \(\mathcal{P}\) and applies alternative convexification strategies. The re-convexification process generates a new convexified problem:
\begin{equation}
    \mathcal{P}_c^{(l)} = \text{ReConvexify}(\mathcal{P}, \{\sigma_l'\}_{l=1}^{k'}, \mathbf{x}^{*(l-1)}),
\end{equation}
where \(\{\sigma_l'\}\) represents alternative convexification strategies. The new problem is solved to obtain \(\mathbf{x}^{*(l)}\).

\end{itemize}

The FDC process terminates when either a feasible solution is found or the maximum iteration count is reached:
\begin{equation}
    \mathbf{x}^{(\text{final})} = \begin{cases}
        \mathbf{x}^{*(l)} & \text{if } \mathbb{I}_f(\mathbf{x}^{*(l)}) = 1 \text{ for some } l \leq L, \\
        \mathbf{x}^{*(L)} & \text{otherwise}.
    \end{cases}
\end{equation}

\section{Experimental Setup}
\subsection{Dataset}
To comprehensively evaluate the performance of NC2C in automated non-convex to convex transformation, we conduct experiments on four diverse datasets that cover different aspects of optimization problem solving. We choose \textbf{NL4Opt} \cite{ramamonjison2023nl4opt}, \textbf{NLP4LP} \cite{ahmaditeshnizi2023optimus}, \textbf{ComplexOR} \cite{xiao2023chain}, \textbf{WireOpt} \cite{peng2025llm} as our experiment datasets. Detailed of the datasets are list in the Appendix~\ref{sec:dataset}.

\begin{table*}[t]
\centering
\caption{
Main experimental results on NL4Opt, NLP4LP, ComplexOR, and NC2C datasets.
We report Success Rate (SR, \%) and Execution Rate (ER, \%) for each dataset.
Results are averaged over 10 independent runs for each problem.
}
\label{tab:main_results}
\resizebox{0.98\textwidth}{!}{
\begin{tabular}{llllllllllll}
\toprule
\multirow{3}{*}{\textbf{Model}} 
& \multirow{3}{*}{\textbf{Method}} 
& \multicolumn{2}{c}{\textbf{NL4Opt}} 
& \multicolumn{2}{c}{\textbf{NLP4LP}} 
& \multicolumn{2}{c}{\textbf{ComplexOR}}
& \multicolumn{2}{c}{\textbf{WireOpt}} \\
\cmidrule(lr){3-4}
\cmidrule(lr){5-6}
\cmidrule(lr){7-8}
\cmidrule(lr){9-10}
& & SR & ER & SR & ER & SR & ER & SR & ER \\
\midrule

\multirow{5}{*}{GPT-5.1}
& Vanilla        
    &  85.0 &  90.2   
    &  78.5 &  85.3   
    &  52.8 &  60.0   
    &  73.0 &  80.0 \\
& Chain-of-Experts        
    &  88.3 &  92.5 
    &  82.0 &  88.2 
    &  58.3 &  66.7 
    &  77.0 &  84.0 \\
& OptiMUS          
    &  90.2 &  94.0 
    &  85.4 &  90.5 
    &  63.9 &  72.2 
    &  80.0 &  87.0 \\
& Reflexion 
    &  87.5 &  92.8 
    &  81.2 &  88.0 
    &  55.6 &  64.4 
    &  75.0 &  83.0 \\
& \cellcolor{red!20}NC2C 
    & \cellcolor{red!20}\textbf{94.5} & \cellcolor{red!20}\textbf{96.7} 
    & \cellcolor{red!20}\textbf{90.5} & \cellcolor{red!20}\textbf{94.2} 
    & \cellcolor{red!20}\textbf{72.2} & \cellcolor{red!20}\textbf{80.6} 
    & \cellcolor{red!20}\textbf{87.0}
    & \cellcolor{red!20}\textbf{93.0} \\
\midrule

\multirow{5}{*}{Qwen3-235B-A22B}
& Vanilla         
    &  78.5 &  85.2   
    &  72.3 &  80.1   
    &  45.0 &  52.8   
    &  68.0 &  75.0 \\
& Chain-of-Experts        
    &  82.3 &  88.5 
    &  76.5 &  83.2 
    &  50.0 &  58.3 
    &  72.0 &  79.0 \\
& OptiMUS          
    &  85.0 &  90.5 
    &  80.2 &  87.0 
    &  55.6 &  63.9 
    &  75.0 &  82.0 \\
& Reflexion 
    &  80.8 &  88.0 
    &  74.5 &  82.5 
    &  48.3 &  56.1 
    &  70.0 &  78.0 \\
& \cellcolor{red!20}NC2C 
    & \cellcolor{red!20}\textbf{91.2} & \cellcolor{red!20}\textbf{94.4} 
    & \cellcolor{red!20}\textbf{86.5} & \cellcolor{red!20}\textbf{91.0} 
    & \cellcolor{red!20}\textbf{66.7} & \cellcolor{red!20}\textbf{75.0} 
    & \cellcolor{red!20}\textbf{82.0}
    & \cellcolor{red!20}\textbf{89.0} \\

\bottomrule
\end{tabular}
}
\end{table*}

\subsection{Baseline Methods}
To comprehensively evaluate NC2C's performance, we compare it against state-of-the-art baseline methods that represent different approaches to optimization problem solving with LLMs, including \textbf{Reflexion}~\cite{shinn2023reflexion}, \textbf{Chain-of-Experts}~\cite{xiao2023chain}, \textbf{OptiMUS}~\cite{ahmaditeshnizi2023optimus}, \textbf{Vanilla LLMs}. More detailed are shown in the Appendix~\ref{sec:baselines}.

\subsection{Evaluation Metrics}
The methods are evaluated based on two primary metrics: {Success Rate} and {Execution Rate}\cite{ahmaditeshnizi2023optimus}. For each optimization problem $\mathcal{P} \in \mathcal{D}$, a total of $N = 10$ evaluations are conducted using both models (GPT-5.1 and Qwen3-235B-A22B), and the rates are computed as the average of these evaluations.

\begin{itemize}
    \item \textbf{Success Rate} is defined as the proportion of outputs that yield an optimal solution based on code execution and lie within the feasible domain while satisfying all constraints, expressed as 
    \begin{equation}
        \text{Success Rate} = \frac{1}{|\mathcal{D}|} \sum_{\mathcal{P} \in \mathcal{D}} \left( \frac{1}{N} \sum_{i=1}^{N} \mathcal{V}_\mathcal{P}\right),
    \end{equation}
    where \(\mathcal{V}_\mathcal{P} = 1\) indicates the solution of problem \(\mathcal{P}\) is feasible, and \(\mathcal{V}_\mathcal{P} = 0\) otherwise.

    \item \textbf{Execution Rate} measures the proportion of generated code that successfully executes and produces outputs, expressed as
    \begin{equation}
            \text{Execution Rate} = \frac{1}{|\mathcal{D}|} \sum_{\mathcal{P} \in \mathcal{D}} \left( \frac{1}{N} \sum_{i=1}^{N} \mathcal{Q}_\mathcal{P} \right),
    \end{equation}
   where \(\mathcal{Q}_\mathcal{P} = 1\) represents the code of problem \(\mathcal{P}\) is successful executed, and \(\mathcal{Q}_\mathcal{P} = 0\) otherwise.
\end{itemize}

\begin{figure}[t]
    \centering
    \includegraphics[width=0.5\textwidth]{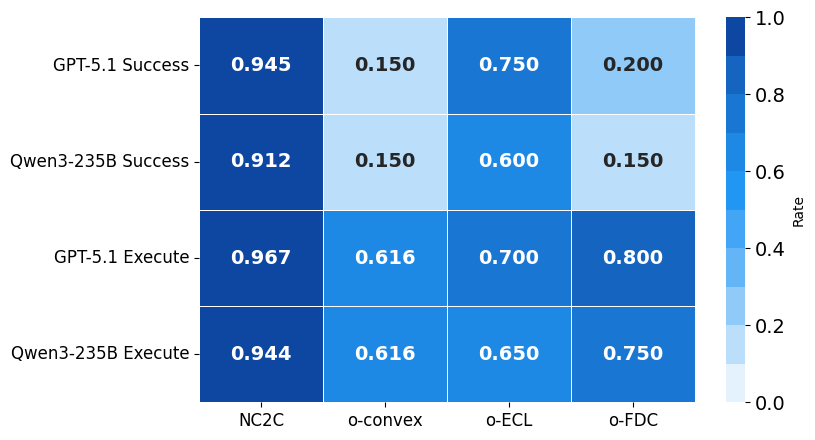}
    \caption{Impact of omitting key components of the NC2C framework on success and execution rates across GPT-5.1 and Qwen3-235B-A22B.}
    \label{fig:heat}
\end{figure}

\begin{figure*}[h]
    \centering
    \hspace{-0.5cm}
\includegraphics[width=1\textwidth]{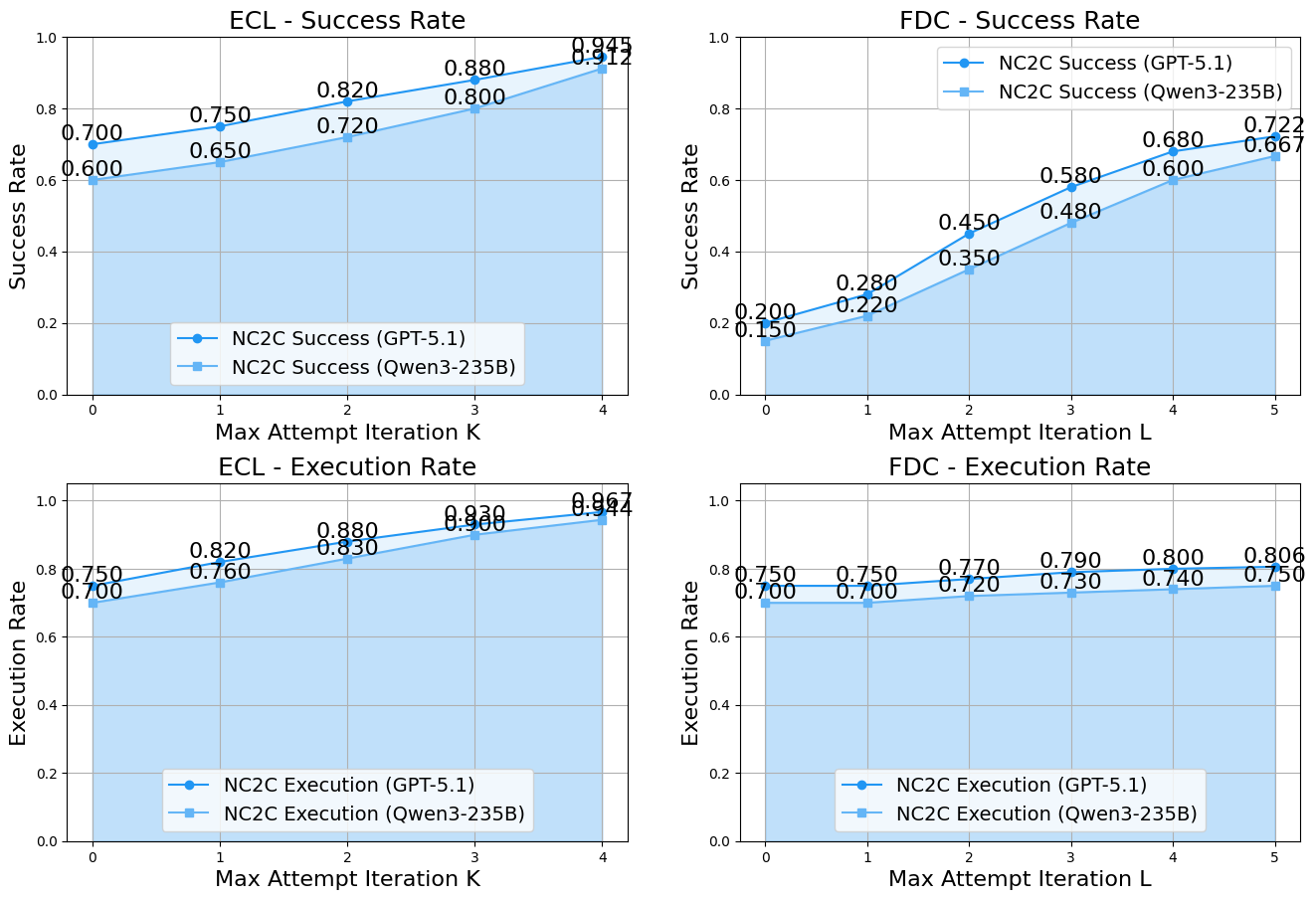}
    \caption{Success and execution rates for NC2C under different maximum iteration counts for ECL and FDC.}
    \label{fig:iteration}
\end{figure*}

\section{Experiment Results}

\subsection{Main Experiments}

We evaluate NC2C and all baseline methods across the four datasets using GPT-5.1 and Qwen3-235B-A22B as the underlying LLMs. For each optimization problem $\mathcal{P} \in \mathcal{D}$, we conduct $N = 10$ independent evaluations, and all reported metrics are averaged over these evaluations. Table~\ref{tab:main_results} presents the comprehensive experimental results.

\textbf{NC2C achieves superior performance across all datasets and models.} As shown in Table~\ref{tab:main_results}, NC2C consistently outperforms all baseline methods on both GPT-5.1 and Qwen3-235B-A22B. On the NL4Opt dataset with GPT-5.1, NC2C achieves a success rate of 94.5\% and execution rate of 96.7\%, significantly surpassing OptiMUS (90.2\% SR, 94.0\% ER), Chain-of-Experts (88.3\% SR, 92.5\% ER), Reflexion (85.0\% SR, 90.2\% ER), and Vanilla (87.5\% SR, 92.8\% ER). On Qwen3-235B-A22B, NC2C achieves 91.2\% success rate and 94.4\% execution rate on NL4Opt, outperforming all baselines. The performance gap is even more pronounced on the more challenging ComplexOR dataset, where NC2C with GPT-5.1 achieves 72.2\% success rate compared to OptiMUS's 63.9\%, and with Qwen3-235B-A22B achieves 66.7\% compared to OptiMUS's 55.6\%, demonstrating NC2C's effectiveness in handling complex non-convex optimization scenarios.

\textbf{NC2C shows robust performance across diverse problem domains.} On the NLP4LP dataset, which contains verbose problem descriptions and complex constraints, NC2C achieves 90.5\% success rate with GPT-5.1 and 86.5\% with Qwen3-235B-A22B, outperforming OptiMUS by 5.1\% and 6.3\%, respectively. On our curated NC2C dataset, which spans multiple domains including mathematical programming, engineering design, and scientific computing, NC2C achieves 87.0\% success rate with GPT-5.1 and 82.0\% with Qwen3-235B-A22B, demonstrating its generalizability across different optimization problem types and model architectures.

\subsection{Ablation Studies}
The NC2C framework consists of multiple key processes, assessing the impact of removing these components on model performance. We will analyze the following scenarios:
\begin{itemize}
\item \textbf{NC2C}: The complete framework proposed in this paper.
\item \textbf{o - convex}: The convexification process is removed, and the model directly handles non - convex mathematical problems.
\item \textbf{o - ECL}: ECL is omitted, and no error correction is performed on erroneous code.
\item \textbf{o - FDC}: FDC is omitted, and no re - solving is performed for solutions that do not satisfy feasibility constraints.
\end{itemize}

\begin{figure*}[t]
    \centering
    \includegraphics[width=\textwidth]{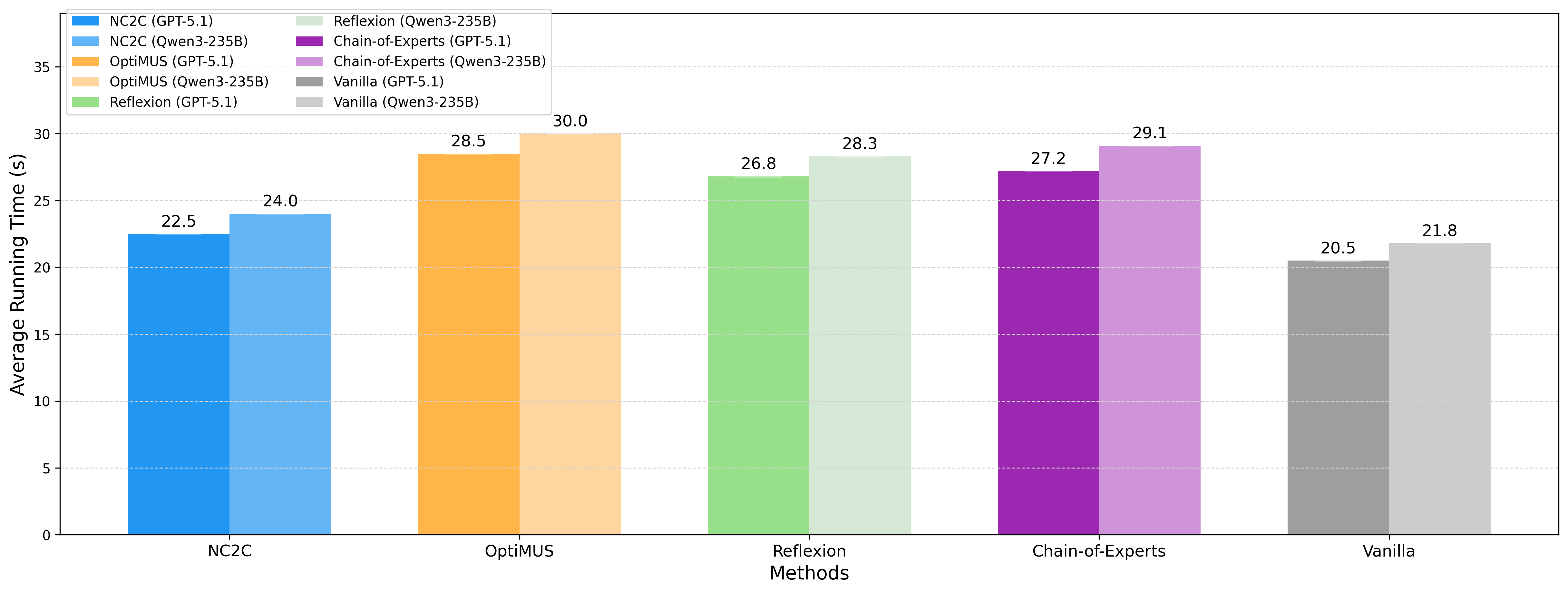}
    \caption{Average Running Time of NC2C and Baseline Methods on GPT-5.1 and Qwen3-235B.}
    \label{fig:time}
\end{figure*}

As shown in the Fig.\ref{fig:heat}, \textbf{o-convex significantly reduces both the success and execution rates, highlighting the importance of convexification within the NC2C framework}.  Specifically, GPT-5.1's success rate drops from 0.945 to 0.15, reflecting an 84.1\% decline, while the execution rate decreases from 0.967 to 0.616. With Qwen3-235B-A22B, the success rate drops from 0.912 to 0.15, reflecting an 83.6\% decline. This highlights the critical need for convexification to transform non-convex problems into solvable forms.

\textbf{{o-ECL negatively impacts the execution rate, which subsequently lowers the success rate.}} For GPT-5.1, the execution rate drops from 0.967 to 0.70, a 27.6\% decrease, and the success rate declines from 0.945 to 0.75, reflecting a 20.6\% reduction. With Qwen3-235B-A22B, the execution rate drops from 0.944 to 0.65, a 31.1\% decrease, and the success rate declines from 0.912 to 0.60, reflecting a 34.2\% reduction. This decrease in execution rates leads to an increase in failed attempts to generate correct solutions, ultimately harming overall success rate.

\textbf{{o-FDC primarily decreases the success rate, emphasizing its role in ensuring solution feasibility.}} Experiments show that without FDC, GPT-5.1's success rate drops to 0.20, a 78.8\% reduction, while Qwen3-235B-A22B's falls to 0.15, an 83.6\% decrease. Although GPT-5.1's execution rate remains at 0.80, the lack of FDC means that many solutions may not be within feasible regions, rendering them ineffective.

\subsection{Maximum Iteration Count Analysis of ECL and FDC}

The previous section highlights the significant roles that ECL and FDC play in the performance of the NC2C framework. Therefore, this section delves deeper into their maximum iteration counts \(K\) and \(L\), respectively, examining their effects on the success and execution rates, as illustrated in Fig.\ref{fig:iteration}.

\textbf{ECL effectively increases success and execution rates in early iterations.} In terms of success rate, ECL quickly identifies and corrects key errors in the early iterations, significantly enhancing overall performance. On the NL4Opt dataset with GPT-5.1, the success rate increases from 70\% to 94.5\%; with Qwen3-235B-A22B, it increases from 60\% to 91.2\%, demonstrating ECL's effectiveness across different model capabilities.

\textbf{FDC demonstrates notable improvements in success rate and maintains steady execution rate enhancements with increasing iterations.} For the success rate on ComplexOR dataset, GPT-5.1 improves from 45\% to 72.2\%, and Qwen3-235B-A22B rises from 35\% to 66.7\%, demonstrating FDC's strong effectiveness in correcting infeasible solutions. This improvement results from FDC's ability to adjust based on the number of iterations \(L\); it provides corrections during initial optimization attempts and reanalyzes the problem after reaching \( \left\lfloor \frac{L}{2} \right\rfloor \). In terms of execution rates, GPT-5.1's execution rate increases from 75\% to 80.6\% on ComplexOR, while Qwen3-235B-A22B rises from 70\% to 75.0\%.

\subsection{Running Time Analysis of NC2C}
To explore whether the running time of NC2C is within an acceptable range while maintaining high success and execution rates, we conduct a comparative experiment on program running times across all baseline methods.

As shown in Fig.\ref{fig:time}, NC2C demonstrates excellent efficiency across different models. With GPT-5.1, NC2C achieves an average running time of 22.5s per problem, which is faster than OptiMUS (28.5s), Reflexion (26.8s), and Chain-of-Experts (27.2s), while only being slightly slower than Vanilla (20.5s). This represents a 21.1\% reduction compared to OptiMUS, a 16.0\% reduction compared to Reflexion, and a 17.3\% reduction compared to Chain-of-Experts. On Qwen3-235B, NC2C also demonstrates superior efficiency. 

Compared with baseline methods, NC2C achieves superior performance with faster running times. While Vanilla requires the least computational time due to its direct approach without additional processing, NC2C's slightly longer running time is justified by its significantly higher success and execution rates. More importantly, NC2C outperforms all other methods in both performance metrics and computational efficiency, demonstrating that its integrated convexification, error correction, and feasibility verification mechanisms are not only effective but also efficiently implemented.

\section{Conclusion}
In this paper, we propose NC2C, a groundbreaking LLM-based framework that automatically transforms generic non-convex optimization problems into convex forms. Our experimental analysis shows that NC2C achieves an execution rate of 89.3\% and a success rate of 76\% on the GPT-4 model, significantly outperforming baseline methods. The framework's success hinges on its integrated components as convexification, error correction, and feasibility checking, each of which plays a critical role in enhancing solution quality and overall robustness. As Language models that combine real-time feedback could achieve more efficient optimization, NC2C sets a new standard for automated non-convex to convex transformation, demonstrating its capabilities in handling diverse optimization problems across various domains. 

\section*{Limitations}

While NC2C demonstrates significant improvements in automated non-convex to convex transformation, several limitations should be acknowledged. The framework's performance is inherently dependent on the underlying LLM's mathematical reasoning capabilities, and its effectiveness may vary with different model architectures or when applied to domains with highly specialized terminology. Although NC2C achieves competitive running times compared to sophisticated baseline methods, the multi-stage processing pipeline introduces computational overhead that may limit its suitability for resource-constrained environments or real-time applications. The convexification process involves relaxation and approximation techniques that may introduce suboptimality, and while NC2C ensures feasibility through the Feasibility Domain Correction mechanism, the gap between the optimal solution of the original non-convex problem and the convexified problem is not always quantifiable. Additionally, the Error Correction Loop may struggle with fundamental mathematical errors, and the framework's decision-making process for selecting convexification strategies may lack transparency, which could affect trust in critical applications.

\section*{Ethical Considerations}

The development and deployment of automated optimization frameworks like NC2C raise several ethical considerations. Users should understand the framework's limitations and not rely solely on automated solutions for critical decisions without appropriate human oversight, particularly in high-stakes scenarios involving resource allocation or decision-making systems. While NC2C operates on mathematical problem formulations, potential biases may arise from the training data used to develop underlying LLMs or from problem selection in evaluation datasets. The framework's reliance on advanced LLMs and computational resources may create barriers to access for researchers with limited resources, and the computational requirements contribute to energy consumption and carbon emissions. When applied to problems involving sensitive data, users must ensure appropriate data handling practices, as the framework's interaction with LLM APIs may involve transmitting problem descriptions to external services. We encourage the research community to develop evaluation metrics that assess not only technical performance but also fairness, transparency, and ethical implications of automated optimization systems.



\bibliography{acl2024}
\bibliographystyle{acl_natbib}

\newpage
\appendix
\section{Broader Impact And Limitations}
\textbf{Broader Impact.}The proposed NC2C method significantly improves success and execution rates in generic non-convex optimization by introducing automated non-convex to convex transformations and multi-stage solution strategies. We believe that by integrating more optimization techniques, such as reinforcement learning and online learning, large-scale models can further enhance their solving capabilities in complex scenarios. In particular in dynamic environments, models that combine real-time feedback could achieve more efficient optimization. Furthermore, the current model can be further integrated with information retrieval and incremental learning techniques to handle more complex tasks with a large amount of noise or uncertainty in practical applications. We believe that the fusion of these technologies will provide high-quality solutions for the broader optimization field, advancing the development and application of automated convexification techniques.

\section{Baselines}
\label{sec:baselines}
\textbf{Reflexion}~\cite{shinn2023reflexion} leverages LLMs' self-reflection capabilities to improve reasoning performance. The method enables the model to critique its own outputs, identify errors, and regenerate corrected formulations through iterative refinement. Reflexion has demonstrated significant improvements in optimization tasks, achieving accuracy improvements from 78.8\% to 92.3\% on the NL4Opt benchmark. We adapt Reflexion for non-convex to convex transformation by incorporating reflection steps that evaluate the correctness of convexification strategies.

\textbf{Chain-of-Experts}~\cite{xiao2023chain} employs a multi-expert architecture where different expert models are chained together to handle complex reasoning tasks. Each expert specializes in different aspects of the problem-solving process, enabling collaborative problem decomposition and solution generation. This approach has shown effectiveness in complex operations research scenarios, achieving 68.8\% success rate on NL4Opt and 40.5\% on ComplexOR datasets. We configure Chain-of-Experts with experts specialized in mathematical modeling, convexification, and code generation.

\textbf{OptiMUS}~\cite{ahmaditeshnizi2023optimus} is an LLM-based agent designed to automatically formulate and solve Mixed Integer Linear Programming (MILP) problems from natural language descriptions. The framework can develop mathematical models, write and debug solver code, and evaluate generated solutions. OptiMUS has demonstrated superior performance, achieving over 12\% improvement on easy datasets and over 8\% on challenging datasets including NLP4LP compared to existing state-of-the-art methods. We extend OptiMUS to handle non-convex problems by incorporating convexification steps.

\textbf{Vanilla} represents the direct application of the underlying the base LLM models (GPT-5.1 \cite{GPT5} or Qwen3-235B-A22B \cite{yang2025qwen3}) to solve optimization problems without additional frameworks or specialized prompting strategies. This baseline evaluates the raw capability of advanced LLMs in handling non-convex to convex transformation tasks, providing a strong comparison point for our framework.

\section{Dataset}
\label{sec:dataset}

\textbf{NL4Opt}~\cite{ramamonjison2023nl4opt} is a benchmark dataset from NL4Opt competition, designed to evaluate the ability to automatically convert natural language descriptions of optimization problems into solver-ready code. The original dataset contains 289 instances, primarily consisting of Linear Programming (LP) and Mixed Integer Linear Programming (MILP) problems. After data cleaning and validation, we use 214 instances that contain well-formed non-convex optimization problems requiring convexification.

\textbf{NLP4LP}~\cite{ahmaditeshnizi2023optimus} is provided by the OptiMUS benchmark, containing 344 linear and integer programming problems with verbose descriptions and multi-dimensional parameters. The dataset is particularly challenging due to its lengthy problem descriptions and complex constraint structures. After filtering for non-convex problems that require transformation, we retain 178 instances for evaluation.

\textbf{ComplexOR}~\cite{xiao2023chain} initially contains 37 challenging optimization problems involving complex operations research scenarios, including combinatorial optimization tasks. This dataset focuses on evaluating LLMs' reasoning and problem-solving capabilities in complex optimization contexts. After data validation, we use 18 instances that involve non-convex components requiring convexification.

\textbf{WireOpt}~\cite{peng2025llm} is specifically designed for non-convex to convex transformation tasks. We extract 50 non-convex optimization problems from authoritative sources including textbooks, research papers, and optimization libraries~\cite{b3,Hou_Shi_Sherali_2014,MATLAB}. These problems span various domains including mathematical programming, engineering design, and scientific computing in a total of 100 instances. This dataset serves as a complementary evaluation set to assess NC2C's performance on diverse non-convex optimization challenges.

\section{Running time of each component within the NC2C framework}
\begin{figure}[h]
    \includegraphics[width=0.5\textwidth]{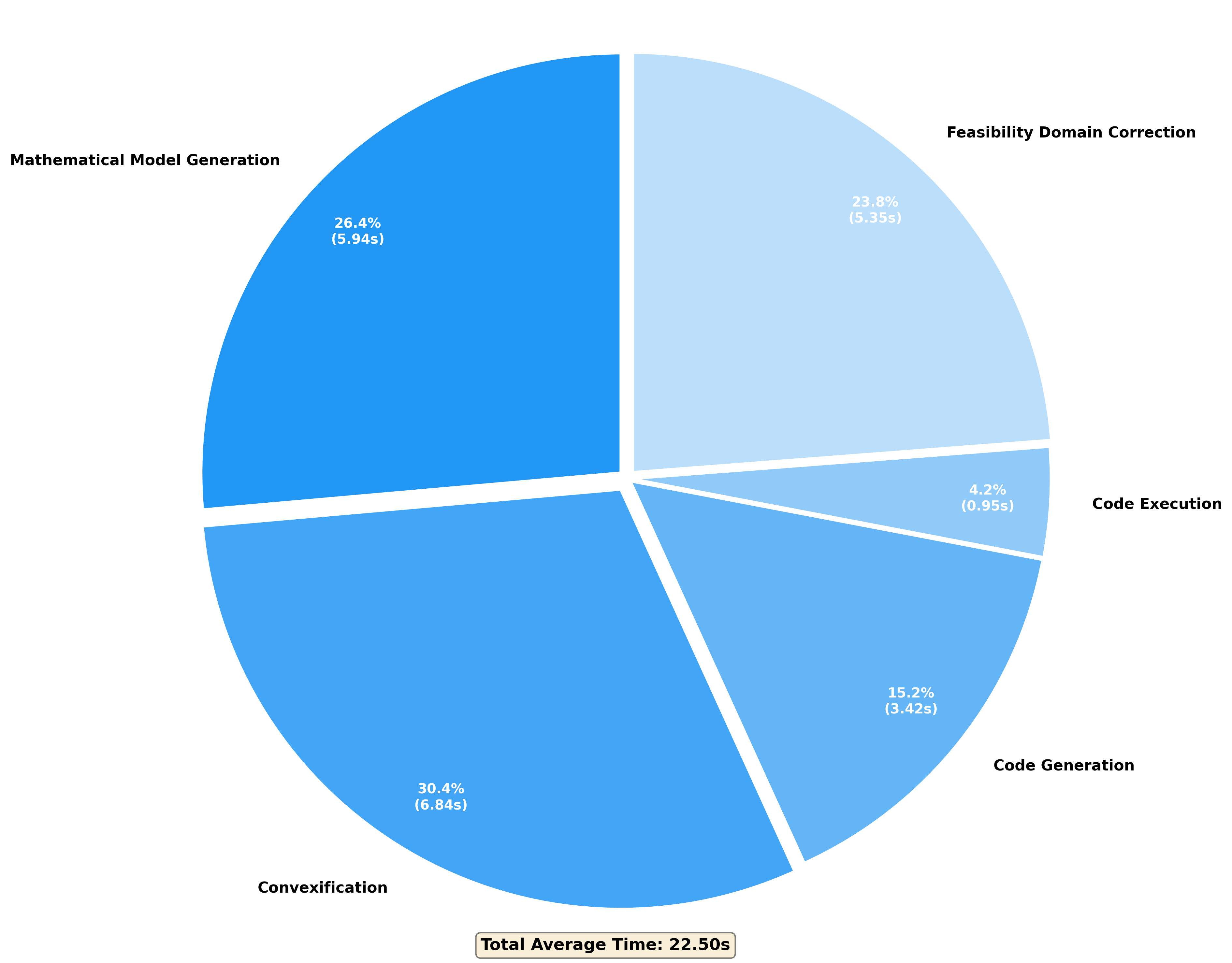}
    \caption{Pie chart of the time proportion of each part.}
    \label{fig:time_bar}
\end{figure}
Building on the running time analysis in last subsection, we continue to analyze the time distribution and contributions of the internal components within the NC2C framework. 

When the NC2C framework runs, convex problem transformation, mathematical model generation, and feasibility verification each take up more than 23\% of the total time. The convex problem transformation part, accounting for about 30.4\% of the time, involves in - depth analysis and conversion of complex non - convex problems into convex ones, entailing numerous complex mathematical operations and logical derivations when handling complex function structures and constraints. The mathematical model generation part, consuming around 26.4\% of the time, demands precise understanding of input problems and strict transformation of natural language descriptions into mathematical models. The pursuit of accuracy in identifying entities, relationships, and semantics in natural language for constructing accurate models leads to significant time investment. The feasibility verification part, taking approximately 23.8\% of the time, has a strict verification mechanism that requires multiple solution checks and possible adjustments. The detailed evaluation of various aspects of the solutions to ensure they meet constraints and practical needs results in increased time consumption. 

As can also be seen from Figure 3 in Section 4 of the main text, in the absence of convex problem transformation (o - convex), the success rate of GPT - 4 plummets from 0.76 to 0.15, a significant drop of 80.3\%, and the execution rate decreases from 0.893 to 0.616. This vividly demonstrates the crucial role of convex problem transformation in converting non - convex problems into solvable forms, and also explains why, despite the significant time consumption, this part is indispensable in the NC2C framework.

Similarly, the high requirements of the mathematical model generation part for understanding and transforming input problems, as well as the strict verification mechanism of the feasibility verification part, result in both of these parts consuming a relatively large amount of time within the framework. However, these components work together. Although they increase the running time, they bring about a performance improvement that far exceeds the time cost for the framework, ensuring the efficiency and reliability of NC2C in solving complex problems. 

In summary, different in time - consumption, these parts collaborate. They achieve performance gains far beyond the time cost, showing the framework's effectiveness and superiority in solving complex non - convex optimization problems.

\section{Pseudocode and Explanation of NC2C Optimization Process}

In the domain of optimization problem - solving, the NC2C optimization process offers an efficient approach. The following pseudocode outlines the core steps of this process, followed by a brief explanation of each significant step.

\onecolumn{
\begin{algorithm}
\caption{NC2C Optimization Process}
\begin{algorithmic}[1]
\State \textbf{Input}: Optimization problem description
\State \textbf{Output}: Success flag, Execute flag, Solution result

\State $\text{success\_flag} \gets 0$
\State $\text{execute\_flag} \gets 0$

\State $\text{math\_formulation} \gets \text{GenerateMathFormulation}(\text{problem})$
\State $\text{convex\_formulation} \gets \text{ConvertToConvex}(\text{math\_formulation})$
\State $\text{code} \gets \text{GenerateCode}(\text{convex\_formulation})$

\State $\text{solution}, \text{execute\_status} \gets \text{ExecuteCode}(\text{code})$
\If {$\text{execute\_status} = \text{Success}$}
    \State $\text{execute\_flag} \gets 1$
    \State $\text{feasibility} \gets \text{CheckFeasibility}(\text{math\_formulation}, \text{solution})$
    \If {$\text{feasibility} = \text{Feasible}$}
        \State $\text{success\_flag} \gets 1$
    \Else
        \For{$\text{attempt} \gets 1 \text{ to } \text{MaxAttempts}$}
            \State $\text{new\_code} \gets \text{RefineCode}(\text{math\_formulation}, \text{code}, \text{attempt}, \text{convex\_formulation})$
            \State $\text{new\_solution}, \text{new\_execute\_status} \gets \text{ExecuteCode}(\text{new\_code})$
            \If {$\text{new\_execute\_status} = \text{Success}$}
                \State $\text{new\_feasibility} \gets \text{CheckFeasibility}(\text{math\_formulation}, \text{new\_solution})$
                \If {$\text{new\_feasibility} = \text{Feasible}$}
                    \State $\text{solution} \gets \text{new\_solution}$
                    \State $\text{success\_flag} \gets 1$
                    \State \textbf{break}
                \EndIf
            \EndIf
        \EndFor
    \EndIf
\EndIf

\State \textbf{Output}: $\text{success\_flag}, \text{execute\_flag}, \text{solution}$
\end{algorithmic}
\end{algorithm}

}

\subsection{Explanation of the Pseudocode}
The entire process aims to solve the optimization problem effectively and obtain a feasible solution, with the status flags indicating the success of different stages of the process.
\begin{enumerate}
    \item \textbf{Initialization}:
        \begin{itemize}
            \item \texttt{success\_flag} and \texttt{execute\_flag} are initialized to 0. These flags will be updated later to indicate the success of the overall process and the code execution respectively.
        \end{itemize}
    \item \textbf{Problem Formulation}:
        \begin{itemize}
            \item \texttt{GenerateMathFormulation(problem)}: Transforms the input natural - language optimization problem into a mathematical formula.
            \item \texttt{ConvertToConvex(math\_formulation)}: Converts the obtained mathematical formula into a convex optimization form, which is easier to solve.
            \item \texttt{GenerateCode(convex\_formulation)}: Generates executable code based on the convex optimization formula.
        \end{itemize}
    \item \textbf{Code Execution and Solution Checking}:
        \begin{itemize}
            \item \texttt{ExecuteCode(code)}: Runs the generated code and returns the solution and the execution status.
            \item \texttt{CheckFeasibility(math\_formulation, solution)}: Checks if the obtained solution is feasible according to the original mathematical formulation.
        \end{itemize}
    \item \textbf{Iterative Refinement}:
        \begin{itemize}
            \item If the initial solution is not feasible, \texttt{RefineCode} is called up to \texttt{MaxAttempts} times to improve the code and try to find a feasible solution.
        \end{itemize}
\end{enumerate}

The effectiveness of these functions \textbf{mainly relies on well - designed prompts}, which will be detailed in the next section of the appendix.

\section{Details of Key Prompt Sections}

In the process of solving optimization problems using LLMs, prompts play a crucial role. They serve as the interface between user problems and model outputs. Well-designed prompts can guide the model to generate expected results. This appendix will elaborate on the key prompts used in each stage from problem input to convex optimization and then to solution. By presenting and analyzing the original text of the prompts, we will reveal their design ideas and working mechanisms.

\subsection{Overall Process Overview}
The entire optimization problem-solving process mainly includes the following key steps:
\begin{enumerate}
    \item \textbf{Problem Input}: Users provide descriptions of actual optimization problems, which will be used as the basic input for subsequent stages.
    \item \textbf{Mathematical Formula Generation}: Utilize carefully designed prompts to guide the LLM to transform the problem described in natural language into a mathematical formula and clarify the optimization objective (maximization or minimization).
    \item \textbf{Convex Optimization Conversion}: Through specific prompts, prompt the LLM to convert the generated non-convex mathematical formula into a convex form for subsequent solution.
    \item \textbf{Code Generation}: Based on the convex optimization formula, use corresponding prompts to let the LLM generate executable code.
    \item \textbf{Code Execution and Solution}: Execute the generated code to obtain the solution of the optimization problem.
\end{enumerate}

\subsection{Details of Prompts in Each Stage}
\subsubsection{Mathematical Formula Generation Stage}
\begin{enumerate}
    \item \textbf{Prompt File Overview}\\
    The main prompt file used in this stage is \texttt{math\_query.txt}. Its core function is to guide the large language model (LLM) to understand and analyze the input optimization problem, generate the corresponding mathematical formula, and clarify the optimization objective. Additionally, it requests the LLM to provide formulas for other variables mentioned in the main formula and their corresponding values.
    \item \textbf{Original Text and Analysis of \texttt{math\_query.txt}}\\
    The original text content of \texttt{math\_query.txt} is as follows:
    \begin{querybox}[title=\texttt{math\_query.txt}]
    Based on this optimization problem, construct a complete mathematical formula, including the objective function and constraints. Please also provide formulas for some of the other variables mentioned in the formula, such as channel condition \(h\), and the values corresponding to these variables.\\
    Input: \$input\$
    \end{querybox}
    Detailed analysis of the original text:
    \begin{itemize}
        \item \textbf{Problem Description Placeholder}: The \texttt{\$input\$} in the original text is a placeholder, which will be replaced by the specific optimization problem description in actual use. This design makes the prompt highly versatile, enabling it to handle different optimization problems. For instance, if the optimization problem is about resource allocation in a network system, the \texttt{\$input\$} will be replaced with the detailed description of this resource - allocation problem, including parameters like available resources, entity demands, etc.
        \item \textbf{Comprehensive Formula Requirement}: The prompt asks the LLM to construct a complete mathematical formula, covering the objective function and constraints. This ensures that the generated mathematical representation fully captures the essence of the optimization problem. Moreover, it specifically requests formulas for other relevant variables (e.g., channel condition \(h\)) and their corresponding values. This detailed requirement helps to build a more accurate and practical mathematical model.
        \item \textbf{Optimization Goal Specification in Code}: Although the original prompt text in \texttt{math\_query.txt} does not explicitly mention the optimization goal specification format, in the Python code `generate\_math`, an additional instruction is added to the user prompt. The line `user\_prompt += "Please specify whether to maximize or minimize, using the format '[Optimization Flag: 1]' for maximize and '[Optimization Flag: 0]' for minimize."` forces the LLM to clearly indicate the optimization objective in the generated mathematical formula. This clear requirement helps the model output results that meet the specifications and facilitates subsequent processing and analysis.
    \end{itemize}
    \item \textbf{Code - based Process Explanation}\\
    The `generate\_math` function implements the process of generating a mathematical formula based on the optimization problem. Here is a step - by - step explanation:
    \begin{itemize}
        \item \textbf{Initialization}: The function initializes an empty string `math` to store the generated mathematical formula and a variable `optimization\_flag` to record the optimization objective.
        \item \textbf{Prompt Preparation}: It reads the content of \texttt{math\_query.txt} as the base user prompt. Then, it replaces the \texttt{\$input\$} placeholder with the actual optimization problem description.
        \item \textbf{Optimization Goal Instruction Addition}: An instruction about specifying the optimization goal is appended to the user prompt.
        \item \textbf{Iterative Generation and Validation}: The function enters a loop where it calls the GPT model using the prepared user prompt to generate a mathematical formula. Then, it constructs a validation prompt to check if the generated formula is consistent with the problem description. If the validation result indicates consistency, the loop is terminated.
        \item \textbf{Optimization Flag Extraction}: After obtaining a valid mathematical formula, the function extracts the optimization flag from the formula. If `[Optimization Flag: 1]` is found, it sets `optimization\_flag` to 1 (maximize); if `[Optimization Flag: 0]` is found, it sets `optimization\_flag` to 0 (minimize).
        \item \textbf{Return Result}: Finally, the function returns the generated mathematical formula and the optimization flag.
    \end{itemize}
    \item \textbf{Example of Generating a Mathematical Formula}\\
    Suppose the optimization problem is: "In a resource allocation problem, we have \(N\) entities. The goal is to maximize the total utility of all entities. The utility of each entity \(i\) is given by \(R_i=\log_2(1 + \frac{p_ih_i}{\sigma^2})\), where \(p_i\) is the resource allocated to entity \(i\), \(h_i\) is the utility gain parameter of entity \(i\), and \(\sigma^2\) is a base constraint parameter. The total available resources is \(P_{total}\), so \(\sum_{i = 1}^{N}p_i\leq P_{total}\), and \(p_i\geq0\) for all \(i\)."
    After replacing the \texttt{\$input\$} in the prompt and adding the optimization goal instruction, the GPT model might generate the following mathematical formula:
    \[
    \max_{p_1,p_2,\cdots,p_N}\sum_{i = 1}^{N}\log_2\left(1+\frac{p_ih_i}{\sigma^2}\right)
    \]
    subject to
    \[
    \begin{cases}
    \sum_{i = 1}^{N}p_i\leq P_{total}\\
    p_i\geq0, \quad i = 1,2,\cdots,N
    \end{cases}
    \]
    and the formula might also contain `[Optimization Flag: 1]` to indicate maximization.
\end{enumerate}

\subsubsection{Convex Optimization Conversion Stage}
\begin{enumerate}
    \item \textbf{Prompt File Overview}\\
    In this stage, two prompt files \texttt{convex\_example.txt} and \texttt{convex\_query.txt} are used. \texttt{convex\_example.txt} provides comprehensive examples of converting non - convex problems into convex problems, serving as references and learning templates for the large language model (LLM). These examples include detailed analyses of non - convex parts, convexification approaches, formula derivations, and the final convex optimization problems. \texttt{convex\_query.txt} contains the input information of the specific non - convex mathematical formula to be convexly converted, guiding the model to perform the conversion operation by specifying clear requirements for the output.
    \item \textbf{Original Text and Analysis of \texttt{convex\_example.txt} and \texttt{convex\_query.txt}}\\
    \begin{itemize}
        \item \textbf{Original Text and Analysis of \texttt{convex\_example.txt}}\\
        The original text content of \texttt{convex\_example.txt} is as follows:
        \begin{codebox}[title=\texttt{convex\_example.txt}]
        Remember your job: 
        1. Following the input and output format, aligning with the length and context.

        Example:

        1:
        Input: 
        \begin{enumerate}
            \item \textbf{Mathematical Formulation: Utility Maximization Problem}
            \item \textbf{Objective Function}
                The goal is to maximize the total utility across all entities:
                \[
                \max_{p, \alpha} \sum_{i} \log\left(1 + \frac{|h_{ii}|^2 p_i}{\sum_{j \neq i} |h_{ij}|^2 p_j + \sigma^2}\right)
                \]
                where:
                \begin{itemize}
                    \item \( p_i \): Resource allocated to entity \( i \) (continuous).
                    \item \( h_{ii} \): Utility gain parameter for entity \( i \).
                    \item \( h_{ij} \): Interaction parameter from entity \( j \) to entity \( i \) (continuous).
                    \item \( \sigma^2 \): Base constraint parameter.
                \end{itemize}
            \item \textbf{Constraints}
                \begin{enumerate}
                    \item \textbf{Resource Constraints}:
                        \begin{enumerate}
                            \item Individual entity resource limits:
                                \[
                                0 \leq p_i \leq P_{\text{max}}, \quad \forall i
                                \]
                            \item Total resource budget:
                                \[
                                \sum_{i} p_i \leq P_{\text{total}}
                                \]
                        \end{enumerate}
                    \item \textbf{Minimum Requirement Constraints}:
                        Each entity must satisfy the minimum requirement:
                        \[
                        \frac{|h_{ii}|^2 p_i}{\sum_{j \neq i} |h_{ij}|^2 p_j + \sigma^2} \geq \gamma_{\text{min}}, \quad \forall i
                        \]
                    \item \textbf{Allocation Constraints}:
                        Each entity can only be assigned to one category:
                        \[
                        \alpha_{ik} \in \{0, 1\}, \quad \forall i, \quad \sum_{k} \alpha_{ik} = 1
                        \]
                \end{enumerate}
            \item \textbf{Additional Variable Definitions}
                \begin{enumerate}
                    \item \textbf{Utility Gain Parameter Definitions}:
                        \begin{enumerate}
                            \item \textbf{Scaling Function}: \( L(d) \), which can be expressed as:
                                \[
                                L(d) = L_0 + 10n \log_{10}\left(\frac{d}{d_0}\right)
                                \]
                                where \( L_0 \) is the reference value at distance \( d_0 \), \( n \) is a scaling exponent, and \( d \) is a distance parameter.
                            \item \textbf{Gain Model}: The utility gain parameters can be modeled as:
                                \[
                                h_{ii} = \sqrt{g_{ii}} L(d_{ii})
                                \]
                                \[
                                h_{ij} = \sqrt{g_{ij}} L(d_{ij})
                                \]
                                where \( g_{ii} \) and \( g_{ij} \) are random variables representing variability, typically modeled as complex Gaussian variables.
                        \end{enumerate}
                \end{enumerate}
        \end{enumerate}
        Output: 
        \begin{enumerate}
            \item \textbf{Identification of Non - Convex Parts}
                The optimization problem presented contains several non - convex components due to the following:
                \begin{enumerate}
                    \item \textbf{Logarithmic Function}: The objective function includes a logarithmic term, which is concave, but the overall maximization creates a non - convex problem.
                    \item \textbf{Ratio Constraints}: The ratio constraints can also introduce non - convexity because they involve ratios of quadratic forms.
                    \item \textbf{Binary Variables}: The use of binary variables \(\alpha_{ik}\) creates a mixed - integer programming (MIP) problem, which is generally non - convex.
                \end{enumerate}
            \item \textbf{Convexification Approach}
                To address the non - convexities, the following strategies can be employed:
                \begin{enumerate}
                    \item \textbf{Change of Variables}: Introduce auxiliary variables to reformulate the ratio constraints into a convex form.
                    \item \textbf{Logarithm Reformulation}: Convert the logarithmic objective to a form suitable for optimization.
                    \item \textbf{Relaxation of Binary Variables}: Use continuous relaxation for binary variables where appropriate.
                \end{enumerate}
            \item \textbf{Derivation for Convexification}
                \begin{enumerate}
                    \item \textbf{Logarithmic Objective Reformulation}:
                        Let \( z_i = \frac{|h_{ii}|^2 p_i}{\sum_{j \neq i} |h_{ij}|^2 p_j + \sigma^2} \). Instead of maximizing \( \log(1 + z_i) \), we can maximize \( z_i \) directly under certain transformations.
                    \item \textbf{Ratio Constraints}:
                        To make the ratio constraints convex, we can rewrite:
                        \[
                        |h_{ii}|^2 p_i \geq \gamma_{\text{min}} \left( \sum_{j \neq i} |h_{ij}|^2 p_j + \sigma^2 \right)
                        \]
                        This can be treated as a convex constraint by rearranging terms to isolate \( p_i \).
                    \item \textbf{Relaxation of Binary Variables}:
                        Instead of using binary variables \( \alpha_{ik} \), introduce continuous variables \( 0 \leq \alpha_{ik} \leq 1 \) such that:
                        \[
                        \sum_{k} \alpha_{ik} = 1 \quad \forall i
                        \]
                \end{enumerate}
            \item \textbf{Convex Optimization Problem}
                After applying these transformations, the new optimization problem can be expressed as follows:
                \begin{enumerate}
                    \item \textbf{Objective Function}
                        Maximize:
                        \[
                        \max_{p, \alpha} \sum_{i} z_i
                        \]
                        where \( z_i \) is defined as:
                        \[
                        z_i = \frac{|h_{ii}|^2 p_i}{\sum_{j \neq i} |h_{ij}|^2 p_j + \sigma^2}
                        \]
                    \item \textbf{Constraints}
                        \begin{enumerate}
                            \item \textbf{Resource Constraints}:
                                \[
                                0 \leq p_i \leq P_{\text{max}}, \quad \forall i
                                \]
                                \[
                                \sum_{i} p_i \leq P_{\text{total}}
                                \]
                            \item \textbf{Convex Ratio Constraints}:
                                \[
                                |h_{ii}|^2 p_i \geq \gamma_{\text{min}} \left( \sum_{j \neq i} |h_{ij}|^2 p_j + \sigma^2 \right), \quad \forall i
                                \]
                            \item \textbf{Relaxed Allocation Constraints}:
                                \[
                                0 \leq \alpha_{ik} \leq 1, \quad \forall i, k
                                \]
                                \[
                                \sum_{k} \alpha_{ik} = 1, \quad \forall i
                                \]
                        \end{enumerate}
                    \item \textbf{Supplementary Variables}
                        \begin{enumerate}
                            \item \textbf{Utility Gain Parameters}:
                                The utility gain parameters can still be defined as:
                                \[
                                h_{ii} = \sqrt{g_{ii}} L(d_{ii}), \quad h_{ij} = \sqrt{g_{ij}} L(d_{ij})
                                \]
                        \end{enumerate}
                \end{enumerate}
        \end{enumerate}
        This reformulated optimization problem is now in a convex form and can be directly addressed using standard convex optimization solvers, allowing for efficient computation and resource allocation.
        \end{codebox}
        The examples in this file provide ideas and methods for convex conversion for the LLM. By demonstrating different types of non - convex problems and their convex forms after conversion, the model can learn common convex conversion techniques such as change of variables, logarithm reformulation, and relaxation of binary variables.
        \item \textbf{Original Text and Analysis of \texttt{convex\_query.txt}}\\
        The original text content of \texttt{convex\_query.txt} is as follows:
        \begin{querybox}[title=\texttt{convex\_query.txt}]
        Please identify if there are any non - convex parts based on the mathematical formula of the optimization problem you built in the previous step and the type of corresponding variables, and if so, choose the algorithm that you think is most appropriate to make all the non - convex components convex (including mixed integer programming problems) until you can solve them directly with the solver. And tell me the derivation of the formula to make it convex, and finally, give me the optimization problem after the convexity, including some supplementary variable formulas.
        Input: \$input\$

        Output:
        \end{querybox}
        The \texttt{\$input\$} in it is also a placeholder, which will be replaced by the actual non - convex mathematical formula. This prompt directly requires the model to perform a comprehensive convex conversion operation on the input non - convex formula, including identifying non - convex parts, choosing appropriate convexification algorithms, providing formula derivations, and presenting the final convex optimization problem.
    \end{itemize}
    \item \textbf{Code Explanation: \texttt{transfer\_convex} Function}
        The \texttt{transfer\_convex} function is designed to convert a non - convex mathematical formula into a convex one using the LLM. Here is a step - by - step breakdown of what the function does:
        \begin{enumerate}
            \item \textbf{Initialization}:
                The function initializes an empty string \texttt{convex} to store the converted convex mathematical formula.
            \item \textbf{Prompt Loading}:
                It reads the system prompt from \texttt{convex\_example.txt} and the user prompt from \texttt{convex\_query.txt} using the \texttt{pure\_set\_prompt} function. These prompts serve as guidelines for the LLM to understand the task and learn from the provided examples.
            \item \textbf{Prompt Modification}:
                The function replaces the \texttt{\$input\$} placeholder in the user prompt with the actual non - convex mathematical formula (\texttt{math}) passed as an argument.
            \item \textbf{Model Call}:
                It calls the LLM using the \texttt{call\_gpt\_from\_sys} function, passing the system prompt and the modified user prompt. The LLM then processes the input and generates the convex version of the mathematical formula.
            \item \textbf{Result Return}:
                Finally, the function returns the converted convex mathematical formula.
        \end{enumerate}
\end{enumerate}
\subsubsection{Convex Optimization Conversion Stage}
\begin{enumerate}
    \item \textbf{Prompt File Overview}\\
In this stage, two prompt files \texttt{convex\_example.txt} and \texttt{convex\_query.txt} are used. \texttt{convex\_example.txt} provides some examples of converting non-convex problems into convex problems, providing references and learning templates for the LLM; \texttt{convex\_query.txt} contains the input information of the specific non-convex mathematical formula to be convexly converted, guiding the model to perform the conversion operation.
    \item \textbf{Original Text and Analysis of \texttt{convex\_example.txt} and \texttt{convex\_query.txt}}\\
\begin{itemize}
    \item \textbf{Original Text and Analysis of \texttt{convex\_example.txt}}\\
The original text content of \texttt{convex\_example.txt} is as follows:
\begin{codebox}[title=\texttt{convex\_example.txt}]
Example 1:
Non-convex problem: \(\min_{x} x^2 - 2x + 1\) (Non-convex part: \(x^2\) may not be convex in some intervals)
Convex conversion: \(\min_{x} (x - 1)^2\) (Converted to a convex function by completing the square)

Example 2:
Non-convex problem: \(\min_{x,y} x^2y + 3xy - 2\) (Non-convex part: \(x^2y\) is non-convex)
Convex conversion: (List the specific conversion method and result here)
\end{codebox}
The examples in this file provide ideas and methods for convex conversion for the LLM. By demonstrating different types of non-convex problems and their convex forms after conversion, the model can learn common convex conversion techniques such as completing the square and variable substitution.
    \item \textbf{Original Text and Analysis of \texttt{convex\_query.txt}}\\
The original text content of \texttt{convex\_query.txt} is as follows:
\begin{querybox}[title=\texttt{convex\_query.txt}]
Please convert the following non-convex mathematical formula into a convex form:

\$input\$
\end{querybox}
The \texttt{\$input\$} in it is also a placeholder, which will be replaced by the actual non-convex mathematical formula. This prompt directly requires the model to perform convex conversion on the input non-convex formula.
\end{itemize}
\end{enumerate}

\subsubsection{Code Generation Stage}
\begin{enumerate}
    \item \textbf{Prompt File Overview}\\
    In the code generation stage, the prompt files \texttt{code\_example.txt} and \texttt{code\_query.txt} are used. \texttt{code\_example.txt} contains some examples of generating code from mathematical formulas, demonstrating the structure and implementation of the code. \texttt{code\_query.txt} provides the convex optimization formula for which code needs to be generated, guiding the large language model to generate the corresponding executable code.
    \item \textbf{Original Text and Analysis of \texttt{code\_example.txt} and \texttt{code\_query.txt}}\\
    \begin{itemize}
        \item \textbf{Original Text of \texttt{code\_example.txt}}\\
        The original text content of \texttt{code\_example.txt} is as follows:
        \begin{codebox}[title=\texttt{code\_example.txt}]
Suppose we have a simple convex optimization problem with the objective of minimizing the objective function \(f(x) = x^2 + 2x + 1\), where \(x\) is a real - valued variable. Here is a code example using Python and the `scipy.optimize` library to solve this problem:
```python
from scipy.optimize import minimize

Define the objective function
def objective(x):
    return x**2 + 2*x + 1

 Initial guess
x0 = 0

Call the optimizer
result = minimize(objective, x0)

 Output the result
if result.success:
    print(f"Optimization succeeded. The optimal solution is: {result.x[0]}, and the optimal value is: {result.fun}")
else:
    print("Optimization failed")
\end{codebox}
\item \textbf{Analysis of \texttt{codeexample.txt}}\
The examples in \texttt{codeexample.txt} provide the large language model with a specific paradigm for translating mathematical formulas into code implementations. It shows how to use a specific Python library (such as scipy.optimize) to solve convex optimization problems. The example details the definition of the objective function, the setting of the initial guess, and the invocation of the optimizer. This helps the model understand the general process of code generation for convex optimization problems, including how to represent mathematical concepts in code, how to choose appropriate libraries and functions, and how to handle the results of the optimization process.
\item \textbf{Original Text of \texttt{codequery.txt}}\
The original text content of \texttt{codequery.txt} is as follows:
\begin{querybox}[title=\texttt{codequery.txt}]
Please generate Python code to solve the following convex optimization problem.
The convex optimization formula is: $input$.
Use appropriate optimization libraries (e.g., scipy.optimize). Clearly define the objective function, constraints (if any), and initial guesses. Provide clear comments in the code to explain the key steps.
\end{querybox}
\item \textbf{Analysis of \texttt{codequery.txt}}\
\texttt{codequery.txt} guides the large language model in code generation in multiple ways. Firstly, it directly specifies the input, which is the convex optimization formula, enabling the model to focus on the specific problem. Secondly, it recommends using appropriate optimization libraries, giving the model a clear direction on where to find the necessary tools for solving the problem. Thirdly, by requiring the clear definition of the objective function, constraints, and initial guesses, it ensures that the generated code is complete and accurate. The requirement for clear comments in the code also helps to make the generated code more understandable and maintainable.
\end{itemize}
\end{enumerate}
\subsubsection{Code Execution Stage}
\begin{enumerate}
\item \textbf{Prompt File Overview}
In the code execution stage, the prompt file \texttt{executecodequery.txt} is used. Its main purpose is to guide the large language model to analyze and handle the execution results of the generated code, including error handling and result extraction.
\item \textbf{Original Text and Analysis of \texttt{executecodequery.txt}}
\begin{itemize}
\item \textbf{Original Text of \texttt{executecodequery.txt}}
The original text content of \texttt{executecodequery.txt} is as follows:
\begin{querybox}[title=\texttt{execute\_code\_query.txt}]
The following is the code generated to solve the convex optimization problem: $code$.
Execute this code. If the execution fails, analyze the error message and suggest possible solutions. If the execution is successful, extract the optimal solution and the optimal value from the output, and present them in a clear format.
\end{querybox}
\item \textbf{Analysis of \texttt{execute\_code\_query.txt}}\
This prompt guides the large language model to perform two main tasks. When the code execution fails, it requires the model to analyze the error message, which helps the model understand common error types in code execution for convex optimization problems, such as incorrect function calls, improper parameter settings, or library import issues. Based on the analysis, the model is expected to suggest possible solutions, which promotes the model's ability to troubleshoot and optimize code. When the code execution is successful, the model is required to extract the optimal solution and the optimal value, which trains the model's ability to understand and process the output of the optimization process.
\end{itemize}
\end{enumerate}
\subsubsection{Feasibility Check Stage}
\begin{enumerate}
\item \textbf{Prompt File Overview}\
In the feasibility check stage, the prompt file \texttt{feasibilitycheckquery.txt} is used. Its function is to guide the large language model to determine whether the obtained solution of the optimization problem is within the feasible region based on the original problem description and the mathematical formula.
\item \textbf{Original Text and Analysis of \texttt{feasibilitycheckquery.txt}}\
\begin{itemize}
\item \textbf{Original Text of \texttt{feasibilitycheckquery.txt}}\
The original text content of \texttt{feasibilitycheckquery.txt} is as follows:
\begin{querybox}[title=\texttt{feasibility\_check\_query.txt}]
The original convex optimization problem description is: $problem_description$.
The corresponding mathematical formula is: $math formula$.
The obtained solution of the optimization problem is: $solution$.
Determine whether the solution is within the feasible region. If it is, return 1; if not, return 0.
\end{querybox}
\item \textbf{Analysis of \texttt{feasibility\_check\_query.txt}}\
This prompt provides the large language model with all the necessary information for a feasibility check, including the problem description, the mathematical formula, and the obtained solution. By asking the model to determine whether the solution is within the feasible region and return a binary result, it forces the model to understand the constraints and conditions in the convex optimization problem and use logical reasoning to make a judgment. This helps to ensure the quality and validity of the final solution.
\end{itemize}
\end{enumerate}

\section{Case Study}
\subsection{Problem}
\subsubsection{Problem Description}

Consider a non-convex optimization problem where the objective is to maximize a utility function 
subject to resource constraints. The problem involves allocating resources to multiple entities 
to maximize the total utility while ensuring that each entity meets a minimum requirement. \\
This optimization problem aims to optimize resource allocation to enhance overall performance 
while maintaining feasibility for all entities.

\subsubsection{Variable Parameters}
1. \textbf{Number of Entities}: \(5\) \\
2. \textbf{Maximum Resource per Entity}: \(30\) units \\
3. \textbf{Minimum Requirement per Entity}: \(10\) units \\
4. \textbf{Utility Function Parameters}: The utility gain is modeled as:
\[
G_{ij}=\frac{1}{(d_{ij})^{\gamma}}\quad (\gamma = 2.2, d_{ij}\text{ is a distance parameter between entity }i\text{ and }j)
\]
5. \textbf{Base Constraint Parameter}: \(N_0 = 1\times10^{-9}\) \\
6. \textbf{Total Available Resources}: \(120\) units

\section{Answer}
\subsection{Mathematical Formulation}
We start by formulating the problem mathematically. Given an optimization problem with the goal of optimizing resource allocation to maximize utility subject to certain constraints.

\subsubsection{Objective Function}
The objective is to maximize the overall system utility, denoted as \(R\):
\[
\text{maximize } R=\sum_{i = 1}^{N}\log_2\left(1+\frac{P_i|h_i|^2}{N_0}\right)
\]
where:
\begin{itemize}
    \item \(N\) is the number of entities.
    \item \(P_i\) is the resource allocated to entity \(i\).
    \item \(h_i\) is the utility gain parameter for entity \(i\).
    \item \(N_0\) is a base constraint parameter.
\end{itemize}

\subsubsection{Constraints}
\begin{enumerate}
    \item \textbf{Total resource constraint}:
\[
\sum_{i = 1}^{N}P_i\leq P_{\text{max}}
\]
where \(P_{\text{max}}\) is the maximum total available resources.
    \item \textbf{Non - negativity constraints for resources}:
\[
P_i\geq0,\quad\forall i = 1,2,\ldots,N
\]
    \item \textbf{Minimum requirement for each entity (optional)}:
\[
R_i\geq R_{\text{min}},\quad\forall i = 1,2,\ldots,N
\]
where \(R_i=\log_2\left(1+\frac{P_i|h_i|^2}{N_0}\right)\) is the utility for entity \(i\) and \(R_{\text{min}}\) is the minimum required utility.
\end{enumerate}

\subsubsection{Variables and Formulas}
\begin{itemize}
    \item \textbf{Utility Gain Parameter \(h\)} is typically modeled as:
\[
h_i=\sqrt{G_i}\cdot\text{exp}(j\theta_i)
\]
where:
\begin{itemize}
    \item \(G_i\) represents the gain parameter for entity \(i\) (can vary based on problem context).
    \item \(\theta_i\) is a phase parameter (uniformly distributed in \([0, 2\pi]\)).
\end{itemize}
    \item \textbf{Base Constraint Parameter \(N_0\)} can be calculated as:
\[
N_0 = kT\cdot B
\]
where:
\begin{itemize}
    \item \(k\) is a constant parameter.
    \item \(T\) is a scaling factor.
    \item \(B\) is a bandwidth or capacity parameter.
\end{itemize}
\end{itemize}

\subsubsection{Optimization Flag}
Since the objective is to maximize utility:
\[
\text{[Optimization Flag: 1]}
\]

This formulation sets up an optimization problem that can be solved using various optimization techniques and algorithms to find the optimal resource allocation \(P_i\) for each entity in the system. The parameters specified in the problem can be adjusted and implemented according to the model being studied.

\subsection{Identification of Non - Convex Parts}
The optimization problem described contains several non - convex components:
\begin{enumerate}
    \item \textbf{Logarithmic Function}: Although the logarithmic function is concave, maximizing the sum of individual utility terms \(R_i=\log_2\left(1+\frac{P_i|h_i|^2}{N_0}\right)\) creates a non - convex problem due to the interactions between different \(P_i\) variables.
    \item \textbf{Minimum Requirement Constraints}: The constraints \(R_i\geq R_{\text{min}}\) involve logarithmic expressions, which can lead to non - convexities, especially when the resource allocation is in the denominator.
    \item \textbf{Resource Constraints}: The total resource constraint coupled with individual non - negativity constraints does not inherently introduce non - convexity but can complicate the optimization landscape.
\end{enumerate}

\subsection{Proposed Algorithm for Convexification}
To address these non - convexities, we can use the following strategies:
\begin{itemize}
    \item \textbf{Change of Variables}: Reformulate the problem using auxiliary variables.
    \item \textbf{Convex Relaxation}: Transform existing logarithmic terms into convex equivalent expressions.
    \item \textbf{Use of Sequential Convex Programming (SCP)}: An iterative approach that optimizes a convex approximation of the original problem at each step.
\end{itemize}

A specific algorithm suitable for this type of problem is the \textbf{Alternating Direction Method of Multipliers (ADMM)}, which can be effective for dealing with non - convex optimization problems involving separable structures.

\subsection{Derivation for Convexification}
\begin{enumerate}
    \item \textbf{Logarithmic Objective Reformulation}:
We can rewrite \(R_i\) as:
\[
R_i=\log_2\left(1+\frac{P_i|h_i|^2}{N_0}\right)
\]
Using a change of variable, let \(z_i = P_i|h_i|^2\). Then we have:
\[
R_i=\log_2\left(1+\frac{z_i}{N_0}\right)
\]
    \item \textbf{Handling Minimum Requirement Constraints}:
Instead of using the logarithmic form for minimum requirements directly, we can reformulate it to express resources:
\[
\frac{P_i|h_i|^2}{N_0}\geq 2^{R_{\text{min}}}- 1
\]
Rearranging gives:
\[
P_i|h_i|^2\geq N_0(2^{R_{\text{min}}}-1)
\]
This can be made into a convex constraint by separating \(P_i\) from the non - linear aspects.
    \item \textbf{Relaxation}:
The resource constraints remain linear since they are in the form of:
\[
\sum_{i = 1}^{N}P_i\leq P_{\text{max}}
\]
\end{enumerate}

\subsection{Convex Optimization Problem}
After applying these transformations and simplifications, the new optimization problem can be expressed as follows:

\subsubsection{Objective Function}
\[
\text{maximize } R=\sum_{i = 1}^{N}z_i,\quad\text{where }z_i=\log_2\left(1+\frac{P_i|h_i|^2}{N_0}\right)
\]

\subsubsection{Constraints}
\begin{enumerate}
    \item \textbf{Total Resource Constraint}:
\[
\sum_{i = 1}^{N}P_i\leq P_{\text{max}}
\]
    \item \textbf{Non - negativity Constraints}:
\[
P_i\geq0,\quad\forall i = 1,2,\ldots,N
\]
    \item \textbf{Convex Minimum Requirement Constraints}:
\[
P_i|h_i|^2\geq N_0(2^{R_{\text{min}}}-1),\quad\forall i = 1,2,\ldots,N
\]
\end{enumerate}

\subsubsection{Supplementary Variable Formulas}
\begin{itemize}
    \item \textbf{Utility Gain Parameter} remains defined as:
\[
h_i=\sqrt{G_i}\cdot\text{exp}(j\theta_i)
\]
    \item \textbf{Base Constraint Parameter} can be expressed as:
\[
N_0 = kT\cdot B
\]
\end{itemize}

This reformulated optimization problem is now in a convex form and can be addressed using standard convex optimization solvers, enabling efficient computation and resource allocation.

\subsection{Code Generation}
The following is a Python code example using `scipy.optimize.minimize` to solve the optimization problem:
\begin{cbox}
[title=\texttt{code}]
   \begin{lstlisting}[language=Python]
import numpy as np
from scipy.optimize import minimize

# Parameters
N = 5  # Number of entities
P_max = 10  # Maximum total resources
R_min = 5  # Minimum requirement
G = np.array([1, 1, 1, 1, 1])  # Utility gain parameters
N0 = 0.001  # Base constraint parameter

# Objective function
def objective(P):
    z = np.log2(1 + (P * G) / N0)  # Calculate z_i
    return -np.sum(z)  # Minimize negative sum for maximization

# Constraints
def constraint_total_power(P):
    return P_max - np.sum(P)

def constraint_qos(P):
    return np.array([P[i] * G[i] - N0 * (2**(R_min) - 1) for i in range(N)])

# Initial guess
P0 = np.ones(N)

# Constraints dictionary
constr = [{'type': 'ineq', 'fun': constraint_total_power},
          {'type': 'ineq', 'fun': constraint_qos}]

# Bounds
bounds = [(0, P_max)] * N  # Each P_i must be between 0 and P_max

# Solve the optimization problem
result = minimize(objective, P0, bounds=bounds, constraints=constr)

# Output results
print("Objective Function Value:", -result.fun)
print("Optimized Resource Allocations (P_i):", result.x)  # Optimized resource allocations
\end{lstlisting} 
\end{cbox}

\subsection{Optimization Results}
The output of the script shows:
\begin{abox}
[title=\texttt{Output}]
    - The value of the objective function is \(54.8325272595287\).

- Optimized Resource Allocations (\(P_i\)) are \([2. 2. 2. 2. 2.]\).
\end{abox}

The initial solution lies within the feasible region.

\end{document}